\pdfoutput=1

\documentclass[10pt,twocolumn,letterpaper]{article}

\usepackage{cvpr}
\usepackage{times}
\usepackage{epsfig}
\usepackage{graphicx}
\usepackage{amsmath}
\usepackage{amssymb}

\usepackage{acronym}
\usepackage{enumitem}
\usepackage[dvipsnames]{xcolor}
\usepackage{multirow}
\usepackage{algorithmic}
\usepackage[ruled]{algorithm2e}
\usepackage{xspace}
\usepackage{balance}
\usepackage{setspace}
\usepackage[pagebackref=true,breaklinks=true,letterpaper=true,colorlinks,bookmarks=false]{hyperref}

\acrodef{ai}[AI]{Artificial Intelligence}
\acrodef{vqa}[VQA]{Visual Question Answering}
\acrodef{rpm}[RPM]{Raven's Progressive Matrices}
\acrodef{pgm}[PGM]{Procedurally Generating Matrices}
\acrodef{wren}[WReN]{Wild Relational Network}
\acrodef{asig}[A-SIG]{Attributed Stochastic Image Grammar}
\acrodef{drt}[DRT]{Dynamic Residual Tree}
\acrodef{wais}[WAIS]{Wechsler Adult Intelligence Scale}

\newcommand{\beforecaption}{\vspace{-9pt}}
\newcommand{\beforetable}{\vspace{-6pt}}
\newcommand{\aftercaption}{\vspace{-12pt}}
\newcommand{\aftertable}{\vspace{-15pt}}
\newcommand\blfootnote[1]{%
  \begingroup
  \renewcommand\thefootnote{}\footnote{#1}%
  \addtocounter{footnote}{-1}%
  \endgroup
}

\makeatletter
\renewcommand{\paragraph}{%
  \@startsection{paragraph}{4}%
  {\z@}{0ex \@plus 0ex \@minus 0ex}{-1em}%
  {\hskip\parindent\normalfont\normalsize\bfseries}%
}
\makeatother

\cvprfinalcopy 

\def\cvprPaperID{2208} 

\ifcvprfinal\fi
\begin{document}

\title{RAVEN: A Dataset for \underline{R}elational and \underline{A}nalogical \underline{V}isual r\underline{E}aso\underline{N}ing}

\author{Chi Zhang$^{\star,1,2}$ \qquad Feng Gao$^{\star,1,2}$ \qquad Baoxiong Jia$^{1}$ \qquad Yixin Zhu$^{1,2}$ \qquad Song-Chun Zhu$^{1,2}$ \\
$^1$ UCLA Center for Vision, Cognition, Learning and Autonomy \\
$^2$ International Center for AI and Robot Autonomy (CARA)\\
{\tt\small \{chi.zhang,f.gao,baoxiongjia,yixin.zhu\}@ucla.edu, sczhu@stat.ucla.edu}
}

\maketitle

\blfootnote{$^\star$ indicates equal contribution.}

\setstretch{0.96}

\begin{abstract}
Dramatic progress has been witnessed in basic vision tasks involving low-level perception, such as object recognition, detection, and tracking. Unfortunately, there is still an enormous performance gap between artificial vision systems and human intelligence in terms of higher-level vision problems, especially ones involving reasoning. Earlier attempts in equipping machines with high-level reasoning have hovered around \ac{vqa}, one typical task associating vision and language understanding. In this work, we propose a new dataset, built in the context of \ac{rpm} and aimed at lifting machine intelligence by associating vision with structural, relational, and analogical reasoning in a hierarchical representation. Unlike previous works in measuring abstract reasoning using \ac{rpm}, we establish a semantic link between vision and reasoning by providing structure representation. This addition enables a new type of abstract reasoning by jointly operating on the structure representation. Machine reasoning ability using modern computer vision is evaluated in this newly proposed dataset. Additionally, we also provide human performance as a reference. Finally, we show consistent improvement across all models by incorporating a simple neural module that combines visual understanding and structure reasoning.
\end{abstract}

\section{Introduction}

\begin{quote}
    The study of vision must therefore include not only the study of how to extract from images \ldots , but also an inquiry into the nature of the \emph{internal representations} by which we \emph{capture} this information and thus make it available as a \textbf{basis} for \emph{decisions about our thoughts and actions}.
    
    \hfill --- David Marr, 1982~\cite{marr1982vision}
\end{quote}

Computer vision has a wide spectrum of tasks. Some computer vision problems are clearly purely visual, ``capturing'' the visual information process; for instance, filters in early vision~\cite{campbell1968application}, primal sketch~\cite{guo2007primal} as the intermediate representation, and Gestalt laws~\cite{kanizsa1979organization} as the perceptual organization. In contrast, some other vision problems have trivialized requirements for perceiving the image, but engage more generalized problem-solving in terms of relational and/or analogical visual reasoning~\cite{holyoak1996mental}. In such cases, the vision component becomes the ``basis for decisions about our thoughts and actions''. 

\begin{figure}[t!]
    \begin{center}
       \includegraphics[width=\linewidth]{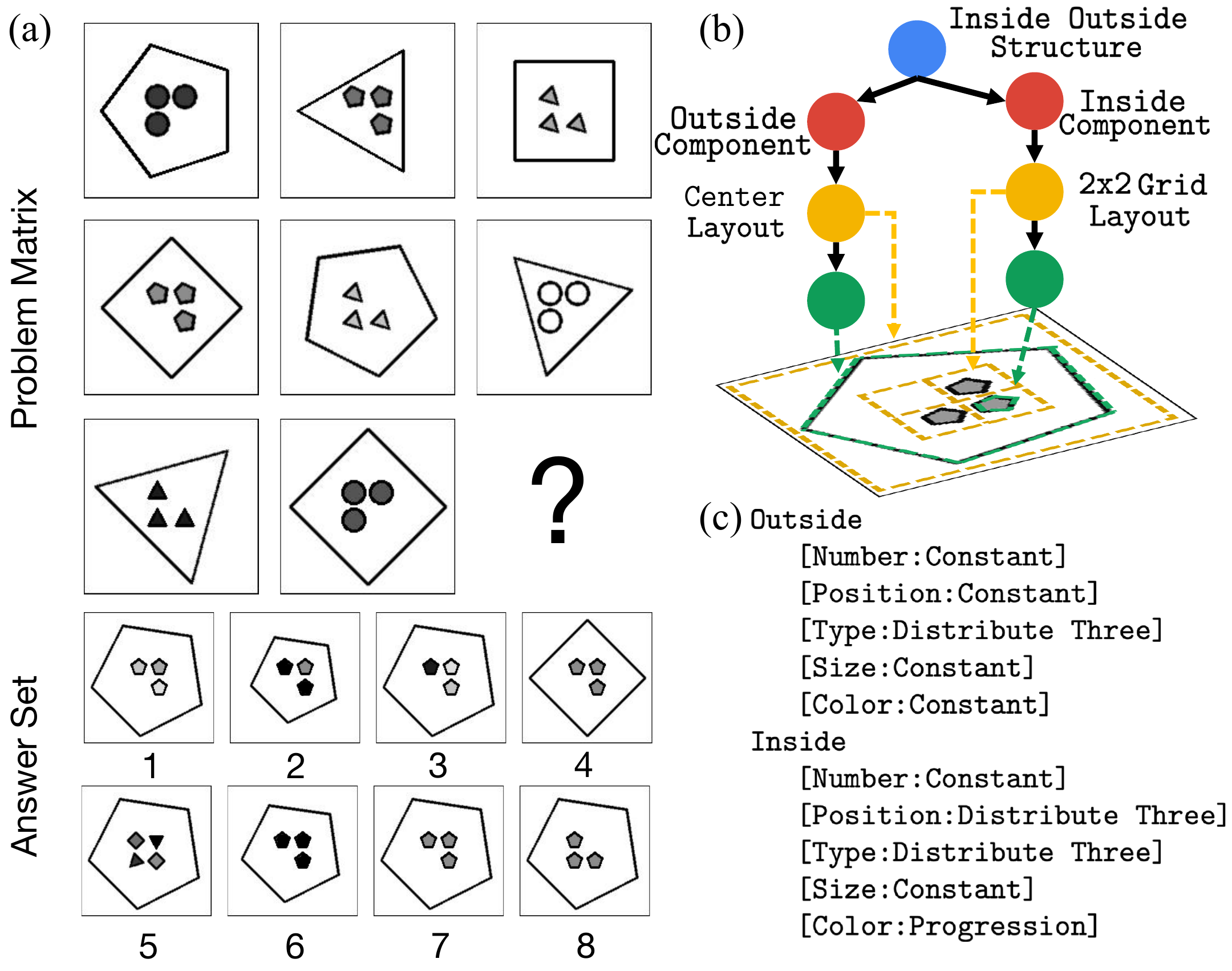}
    \end{center}
    \beforecaption
    \caption{(a) An example \ac{rpm}. One is asked to select an image that best completes the problem matrix, following the structural and analogical relations. Each image has an underlying structure. (b) Specifically in this problem, it is an inside-outside {\color{NavyBlue}{\textbf{structure}}} in which the outside {\color{Red}{\textbf{component}}} is a {\color{YellowOrange}{\textbf{layout}}} with a single centered object and the inside {\color{Red}{\textbf{component}}} is a $2 \times 2$ grid {\color{YellowOrange}{\textbf{layout}}}. Details in Figure~\ref{fig:process}. (c) lists the rules for (a). The compositional nature of the rules makes this problem a difficult one. The correct answer is 7.}
    \label{fig:prologue}
    \aftercaption
\end{figure}


Currently, the majority of the computer vision tasks focus on ``capturing'' the visual information process; few lines of work focus on the later part---the relational and/or analogical visual reasoning. One existing line of work in equipping artificial systems with reasoning ability hovers around \acf{vqa}~\cite{antol2015vqa,johnson2017clevr,ren2015exploring,yi2018neural,zhu2016visual7w}.
However, the reasoning skills required in \ac{vqa} lie only at the periphery of the cognitive ability test circle~\cite{carpenter1990one}. To push the limit of computer vision or more broadly speaking, \acf{ai}, towards the center of cognitive ability test circle, we need a test originally designed for measuring human's intelligence to challenge, debug, and improve the current artificial systems.

A surprisingly effective ability test of human visual reasoning has been developed and identified as the \acf{rpm}~\cite{kunda2013computational,raven1938raven,strannegaard2013anthropomorphic}, which is widely accepted and believed to be highly correlated with real intelligence~\cite{carpenter1990one}. Unlike \ac{vqa}, \ac{rpm} lies directly at the center of human intelligence~\cite{carpenter1990one}, is diagnostic of abstract and structural reasoning ability~\cite{snow1984the}, and characterizes the defining feature of high-level intelligence, \ie, \emph{fluid intelligence}~\cite{jaeggi2008improving}.

Figure~\ref{fig:prologue} shows an example of \ac{rpm} problem together with its structure representation. Provided two rows of figures consisting of visually simple elements, one must efficiently derive the correct image structure (Figure \ref{fig:prologue}(b)) and the underlying rules (Figure~\ref{fig:prologue}(c)) to jointly reason about a candidate image that best completes the problem matrix. In terms of levels of reasoning required, \ac{rpm} is arguably harder compared to \ac{vqa}:
\begin{itemize}[leftmargin=*,noitemsep,nolistsep]
    \item Unlike \ac{vqa} where natural language questions usually imply what to pay attention to in the image, \ac{rpm} relies merely on visual clues provided in the matrix and the \emph{correspondence problem} itself, \ie, finding the correct level of attributes to encode, is already a major factor distinguishing populations of different intelligence~\cite{carpenter1990one}. 
    
    \item While \ac{vqa} only requires spatial and semantic understanding, \ac{rpm} needs joint spatial-temporal reasoning in the problem matrix and the answer set. The limit of \emph{short-term memory}, the ability of \emph{analogy}, and the discovery of the \emph{structure} have to be taken into consideration. 
    
    \item Structures in \ac{rpm} make the compositions of rules much more complicated. Unlike \ac{vqa} whose questions only encode relatively simple first-order reasoning, \ac{rpm} usually includes more sophisticated logic, even with recursions. By composing different rules at various levels, the reasoning progress can be extremely difficult.
\end{itemize}

To push the limit of current vision systems' reasoning ability, we generate a new dataset to promote further research in this area. We refer to this dataset as the Relational and Analogical Visual rEasoNing dataset (RAVEN) in homage to John Raven for the pioneering work in the creation of the original \ac{rpm}~\cite{raven1938raven}. In summary:
\begin{itemize}[leftmargin=*,noitemsep,nolistsep]
    \item RAVEN consists of $1,120,000$ images and $70,000$ \ac{rpm} problems, equally distributed in $7$ distinct figure configurations.
    \item Each problem has $16$ tree-structure annotations, totaling up to $1,120,000$ structural labels in the entire dataset.
    \item We design $5$ rule-governing attributes and $2$ noise attributes. Each rule-governing attribute goes over one of $4$ rules, and objects in the same component share the same set of rules, making in total $440,000$ rule annotations and an average of $6.29$ rules per problem.
\end{itemize}

The RAVEN dataset is designed inherently to be light in visual recognition and heavy in reasoning. Each image only contains a limited set of simple gray-scale objects with clear-cut boundaries and no occlusion. In the meantime, rules are applied row-wise, and there could be one rule for each attribute, attacking visual systems' major weaknesses in \emph{short-term memory} and \emph{compositional reasoning}~\cite{johnson2017clevr}.

An obvious paradox is: in this innately compositional and structured \ac{rpm} problem, no annotations of structures are available in previous works (\eg,~\cite{barrett2018measuring,wang2015automatic}). Hence, we set out to establish a semantic link between visual reasoning and structure reasoning in \ac{rpm}. We ground each problem instance to a sentence derived from an \acf{asig}~\cite{fu1974syntactic,lin2009stochastic,park2015attributed,wu2007compositional,zhu2016reconfigurable,zhu2007stochastic} and decompose the data generation process into two stages: the first stage samples a sentence from a pre-defined \ac{asig} and the second stage renders an image based on the sentence. This structured design makes the dataset very diverse and easily extendable, enabling generalization tests in different figure configurations. More importantly, the data generation pipeline naturally provides us with abundant dense annotations, especially the structure in the image space. This semantic link between vision and structure representation opens new possibilities by breaking down the problem into image understanding and tree- or graph-level reasoning~\cite{kipf2016semi,tai2015improved}. As shown in Section~\ref{sec:experiments}, we empirically demonstrate that models with a simple structure reasoning module to incorporate both vision-level understanding and structure-level reasoning would notably improve their performance in \ac{rpm}.

The organization of the paper is as follows. In Section~\ref{sec:related}, we discuss related work in visual reasoning and computational efforts in \ac{rpm}. Section~\ref{sec:raven} is devoted to a detailed description of the RAVEN dataset generation process, with Section~\ref{sec:analysis} benchmarking human performance and comparing RAVEN with a previous \ac{rpm} dataset. In Section~\ref{sec:model}, we propose a simple extension to existing models that incorporates vision understanding and structure reasoning. All baseline models and the proposed extensions are evaluated in Section~\ref{sec:experiments}. The notable gap between human subjects (84\%) and vision systems (59\%) calls for further research into this problem. We hope RAVEN could contribute to the long-standing effort in human-level reasoning \ac{ai}.

\begin{figure*}[t!]
    \begin{center}
        \includegraphics[width=\linewidth]{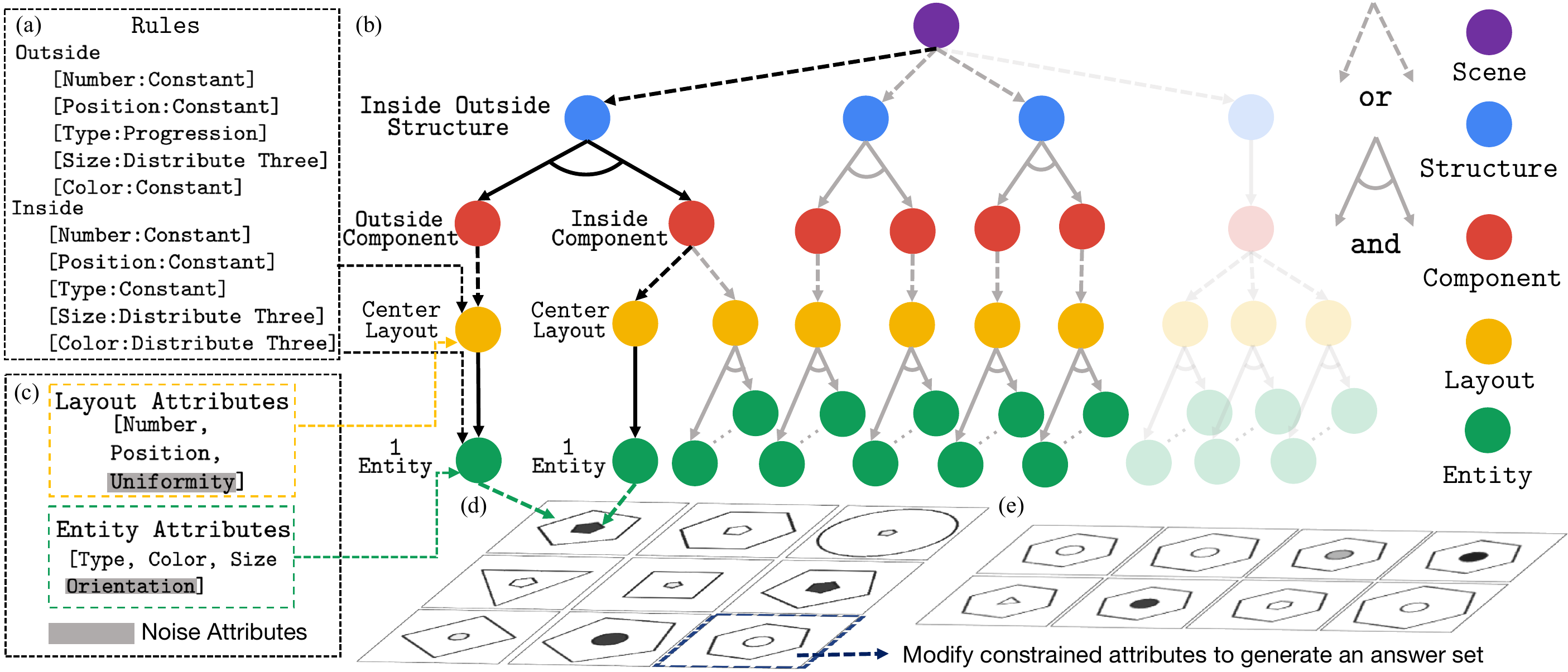}
    \end{center}
    \beforecaption
    \caption{RAVEN creation process. A graphical illustration of the grammar production rules used in \ac{asig} is shown in (b). Note that \texttt{Layout} and \texttt{Entity} have associated attributes (c). Given a randomly sampled rule combination (a), we first prune the grammar tree (the transparent branch is pruned). We then sample an image structure together with the values of the attributes from (b), denoted by black, and apply the rule set (a) to generate a single row. Repeating the process three times yields the entire problem matrix in (d). (e) Finally, we sample constrained attributes and vary them in the correct answer to break the rules and obtain the candidate answer set.}
    \label{fig:process}
    \aftercaption
\end{figure*}

\section{Related Work}\label{sec:related}

\paragraph{Visual Reasoning}
Early attempts were made in 1940s-1970s in the field of logic-based \ac{ai}. Newell argued that one of the potential solutions to \ac{ai} was ``to construct a single program that would take a standard intelligence test''~\cite{newell1973you}. There are two important trials: (i) Evans presented an \ac{ai} algorithm that solved a type of geometric analogy tasks in the \ac{wais} test~\cite{evans1962heuristic,evans1964heuristic}, and (ii) Simon and Kotovsky devised a program that solved Thurstone letter series completion problems~\cite{thurstone1941factorial}. However, these early attempts were heuristic-based with hand-crafted rules, making it difficult to apply to other problems.

\setstretch{0.94}

The reasoning ability of modern vision systems was first systematically analyzed in the CLEVR dataset~\cite{johnson2017clevr}. By carefully controlling inductive bias and slicing the vision systems' reasoning ability into several axes, Johnson \etal successfully identified major drawbacks of existing models. A subsequent work~\cite{johnson2017inferring} on this dataset achieved good performance by introducing a program generator in a structured space and combining it with a program execution engine. A similar work that also leveraged language-guided structured reasoning was proposed in~\cite{hu2017learning}. Modules with special attention mechanism were latter proposed in an end-to-end manner to solve this visual reasoning task~\cite{hudson2018compositional,santoro2017simple,zhu2017structured}. However, superior performance gain was observed in very recent works~\cite{cao2018visual,mascharka2018transparency,yi2018neural} that fell back to structured representations by using primitives, dependency trees, or logic. These works also inspire us to incorporate structure information into solving the \ac{rpm} problem.

More generally, Bisk \etal~\cite{bisk2018learning} studied visual reasoning in a 3D block world. Perez \etal~\cite{perez2018film} introduced a conditional layer for visual reasoning. Aditya \etal~\cite{aditya2018explicit} proposed a probabilistic soft logic in an attention module to increase model interpretability. And Barrett \etal~\cite{barrett2018measuring} measured abstract reasoning in neural networks. 

\paragraph{Computational Efforts in \ac{rpm}}
The research community of cognitive science has tried to attack the problem of \ac{rpm} with computational models earlier than the computer science community. However, an oversimplified assumption was usually made in the experiments that the computer programs had access to a symbolic representation of the image and the operations of rules~\cite{carpenter1990one,lovett2017modeling,lovett2010structure,lovett2009solving}. As reported in Section~\ref{sec:search}, we show that giving this critical information essentially turns it into a searching problem. Combining it with a simple heuristics provides us an optimal solver, easily surpassing human performance. Another stream of \ac{ai} research~\cite{little2012bayesian,mcgreggor2014confident,mcgreggor2014fractals,mekik2018similarity,Shegheva2018TheSA} tries to solve \ac{rpm} by various measurements of image similarity. To promote fair comparison between computer programs and human subjects in a data-driven manner, Wang and Su~\cite{wang2015automatic} first proposed a systematic way of automatically generating \ac{rpm} using first-order logic. Barrett \etal~\cite{barrett2018measuring} extended their work and introduced the \ac{pgm} dataset by instantiating each rule with a relation-object-attribute tuple. Hoshen and Werman~\cite{hoshen2017iq} first trained a CNN to complete the rows in a simplistic evaluation environment, while Barrett \etal~\cite{barrett2018measuring} used an advanced \ac{wren} and studied its generalization.


\section{Creating RAVEN}\label{sec:raven}

Our work is built on prior work aforementioned. We implement all relations in Advanced Raven's Progressive Matrices identified by Carpenter \etal~\cite{carpenter1990one} and generate the answer set following \emph{the monotonicity of \ac{rpm}'s constraints} proposed by Wang and Su~\cite{wang2015automatic}.

Figure~\ref{fig:process} shows the major components of the generation process. Specifically, we use the \ac{asig} as the representation of \ac{rpm}; each \ac{rpm} is a parse tree that instantiates from the \ac{asig}. After rules are sampled, we prune the grammar to make sure the relations could be applied on any sentence sampled from it. We then sample a sentence from the pruned grammar, where rules are applied to produce a valid row. Repeating such a process three times yields a problem matrix. To generate the answer set, we modify attributes on the correct answer such that the relationships are broken. Finally, the structured presentation is fed into a rendering engine to generate images. We elaborate the details below\footnote{See the supplementary material for production rules, semantic meanings of rules and nodes, and more examples.}. 

\begin{figure}[t!]
    \begin{center}
        \includegraphics[width=\linewidth]{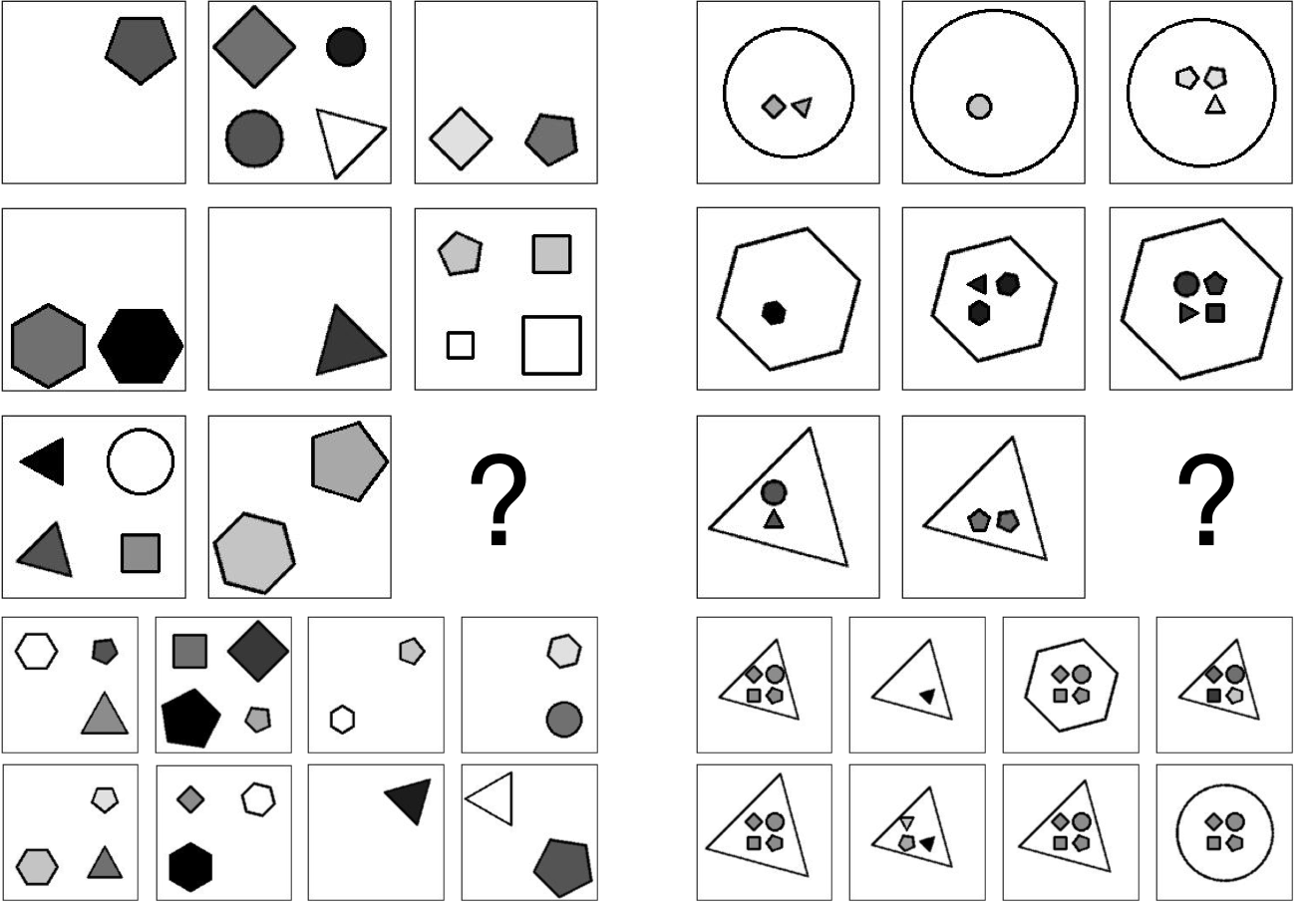}
    \end{center}
    \beforecaption
    \caption{Examples of \ac{rpm} that show the effects of adding \emph{noise} attributes. (Left) \texttt{Position}, \texttt{Type}, \texttt{Size}, and \texttt{Color} could vary freely as long as \texttt{Number} follows the rule. (Right) \texttt{Position} and \texttt{Type} in the inside group could vary freely.}
    \label{fig:noise_attr}
    \aftercaption
\end{figure}

\subsection{Defining the Attributed Grammar}\label{subsec:grammar}

We adopt an \ac{asig} as the hierarchical and structured image grammar to represent the \ac{rpm} problem. Such representation is advanced compared with prior work (\eg,~\cite{barrett2018measuring,wang2015automatic}) which, at best, only maintains a flat representation of rules.

See Figure \ref{fig:process} for a graphical illustration of the grammar production rules. Specifically, the \ac{asig} for \ac{rpm} has $5$ levels---\texttt{Scene}, \texttt{Structure}, \texttt{Component}, \texttt{Layout}, and \texttt{Entity}. Note that each grammar level could have multiple instantiations, \ie, different categories or types. The \texttt{Scene} level could choose any available \texttt{Structure}, which consists of possibly multiple \texttt{Components}. Each \texttt{Component} branches into \texttt{Layouts} that links \texttt{Entities}. Attributes are appended to certain levels; for instance, (i) \texttt{Number} and \texttt{Position} are associated with \texttt{Layout}, and (ii) \texttt{Type}, \texttt{Size}, and \texttt{Color} are associated with \texttt{Entity}. Each attribute could take a value from a finite set. During sampling, both image structure and attribute values are sampled.

To increase the challenges and difficulties in the RAVEN dataset, we further append $2$ types of \emph{noise} attributes---\texttt{Uniformity} and \texttt{Orientation}---to \texttt{Layout} and \texttt{Entity}, respectively. \texttt{Uniformity}, set false, will not constrain \texttt{Entities} in a \texttt{Layout} to look the same, while \texttt{Orientation} allows an \texttt{Entity} to self-rotate. See Figure~\ref{fig:noise_attr} for the effects of the noise attributes.

\begin{figure*}[t!]
    \begin{center}
        \includegraphics[width=\linewidth]{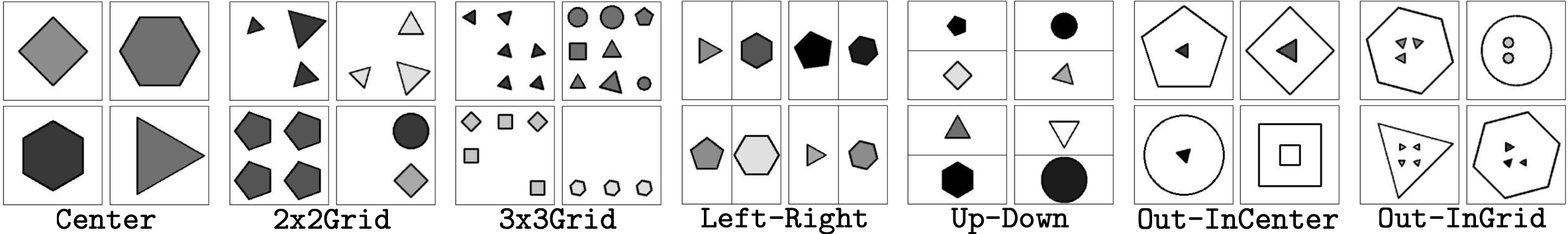}
    \end{center}
    \beforecaption
    \caption{Examples of 7 different figure configurations in the proposed RAVEN dataset.}
    \label{fig:peek_view}
    \aftercaption
\end{figure*}

This grammatical design of the image space allows the dataset to be very diverse and easily extendable. In this dataset, we manage to derive $7$ configurations by combining different \texttt{Structures}, \texttt{Components}, and \texttt{Layouts}. Figure~\ref{fig:peek_view} shows examples in each figure configuration.

\setstretch{0.975}

\subsection{Applying Rules}\label{subsec:relation}

Carpenter \etal~\cite{carpenter1990one} summarized that in the advanced \ac{rpm}, rules were applied row-wise and could be grouped into $5$ types. Unlike Berrett \etal~\cite{barrett2018measuring}, we strictly follow Carpenter \etal's description of \ac{rpm} and implement all the rules, except that we merge \texttt{Distribute Two} into \texttt{Distribute Three}, as the former is essentially the latter with a null value in one of the attributes.

Specifically, we implement $4$ types of rules in RAVEN: \texttt{Constant}, \texttt{Progression}, \texttt{Arithmetic}, and \texttt{Distribute Three}. Different from~\cite{barrett2018measuring}, we add internal parameters to certain rules (\eg, \texttt{Progression} could have increments or decrements of $1$ or $2$), resulting in a total of $8$ distinct rule instantiations. Rules do not operate on the $2$ noise attributes. As shown in Figure~\ref{fig:prologue} and \ref{fig:process}, they are denoted as \texttt{[attribute:rule]} pairs.

To make the image space even more structured, we require each attribute to go over one rule and all \texttt{Entities} in the same \texttt{Component} to share the same set of rules, while different \texttt{Components} could vary.

Given the tree representation and the rules, we first prune the grammar tree such that all sub-trees satisfy the constraints imposed by the relations. We then sample from the tree and apply the rules to compose a row. Iterating the process three times yields a problem matrix.

\subsection{Generating the Answer Set}

To generate the answer set, we first derive the correct representation of the solution and then leverage the monotonicity of \ac{rpm} constraints proposed by Wang and Su~\cite{wang2015automatic}. To break the correct relationships, we find an attribute that is constrained by a rule as described in Section~\ref{subsec:relation} and vary it. By modifying only one attribute, we could greatly reduce the computation. Such modification also increases the difficulty of the problem, as it requires attention to subtle difference to tell an incorrect candidate from the correct one.

\section{Comparison and Analysis}\label{sec:analysis}

In this section, we compare RAVEN with the existing \ac{pgm}, presenting its key features and some statistics in Section~\ref{sec:comparison}. In addition, we fill in two missing pieces in a desirable \ac{rpm} dataset, \ie, structure and hierarchy (Section~\ref{sec:structure}), as well as the human performance (Section~\ref{sec:human}). We also show that \ac{rpm} becomes trivial and could be solved instantly using a heuristics-based searching method (Section~\ref{sec:search}), given a symbolic representation of images and operations of rules.

\subsection{Comparison with \ac{pgm}}\label{sec:comparison}

Table~\ref{tbl:compare} summarizes several essential metrics of RAVEN and \ac{pgm}. Although \ac{pgm} is larger than RAVEN in terms of size, it is very limited in the average number of rules (\textbf{AvgRule}), rule instantiations (\textbf{RuleIns}), number of structures (\textbf{Struct}), and figure configurations (\textbf{FigConfig}). This contrast in \ac{pgm}'s gigantic size and limited diversity might disguise model fitting as a misleading reasoning ability, which is unlikely to generalize to other scenarios.

\begin{table}[hb!]
    \begin{center}
    \beforetable
        \caption{Comparison with the \ac{pgm} dataset.}\label{tbl:compare}
        \begin{tabular}{c c c}
            \hline
                                & \textbf{\ac{pgm}}~\cite{barrett2018measuring}  & \textbf{RAVEN (Ours)} \\
            \hline\hline
            \textbf{AvgRule}    & 1.37      & 6.29       \\
            \textbf{RuleIns}    & 5         & 8          \\
            \textbf{Struct}     & 1         & 4          \\
            \textbf{FigConfig}  & 3         & 7          \\
            \textbf{StructAnno} & 0         & 1,120,000  \\
            \textbf{HumanPerf}  &           & \checkmark \\
            \hline
        \end{tabular}
    \end{center}
    \aftertable
\end{table}

To avoid such an undesirable effect, we refrain from generating a dataset too large, even though our structured representation allows generation of a combinatorial number of problems. Rather, we set out to incorporate more rule instantiations ($8$), structures ($4$), and figure configurations ($7$) to make the dataset diverse (see Figure~\ref{fig:peek_view} for examples). Note that an equal number of images for each figure configuration is generated in the RAVEN dataset.

\subsection{Introduction of Structure}\label{sec:structure}

A distinctive feature of RAVEN is the introduction of the structural representation of the image space. Wang and Su~\cite{wang2015automatic} and Barrett \etal~\cite{barrett2018measuring} used plain logic and flat rule representations, respectively, resulting in no base of the structure to perform reasoning on. In contrast, we have in total $1,120,000$ structure annotations (\textbf{StructAnno}) in the form of parsed sentences in the dataset, pairing each problem instance with $16$ sentences for both the matrix and the answer set. These representations derived from the \ac{asig} allow a new form of reasoning, \ie, one that combines visual understanding and structure reasoning. As shown in~\cite{lovett2017modeling,lovett2010structure,lovett2009solving} and our experiments in Section~\ref{sec:experiments}, incorporating structure into \ac{rpm} problem solving could result in further performance improvement across different models. 

\subsection{Human Performance Analysis}\label{sec:human}

Another missing point in the previous work~\cite{barrett2018measuring} is the evaluation of human performance. To fill in the missing piece, we recruit human subjects consisting of college students from a subject pool maintained by the Department of Psychology to test their performance on a subset of representative samples in the dataset. In the experiments, human subjects were familiarized by solving problems with only one non-\texttt{Constant} rule in a fixed configuration. After the familiarization, subjects were asked to answer \ac{rpm} problems with complex rule combinations, and their answers were recorded. Note that we deliberately included all figure configurations to measure generalization in the human performance and only ``easily perceptible'' examples were used in case certain subjects might have impaired perception. The results are reported in Table~\ref{tbl:plain_training}. The notable performance gap calls for further research into this problem. See Section~\ref{sec:experiments} for detailed analysis and comparisons with vision models.

\subsection{Heuristics-based Solver using Searching}\label{sec:search}

We also find that the \ac{rpm} could be essentially turned into a searching problem, given the symbolic representation of images and the access to rule operations as in~\cite{lovett2017modeling,lovett2010structure,lovett2009solving}. Under such a setting, we could treat this problem as constraint satisfaction and develop a heuristics-based solver. The solver checks the number of satisfied constraints in each candidate answer and selects one with the highest score, resulting in perfect performance. Results are reported in Table~\ref{tbl:plain_training}. The optimality of the heuristic-based solver also verifies the well-formedness of RAVEN in the sense that there exists only one candidate that satisfies all constraints.

\section{Dynamic Residual Tree for \ac{rpm}}\label{sec:model}

\begin{figure}[t!]
    \begin{center}
        \includegraphics[width=\linewidth]{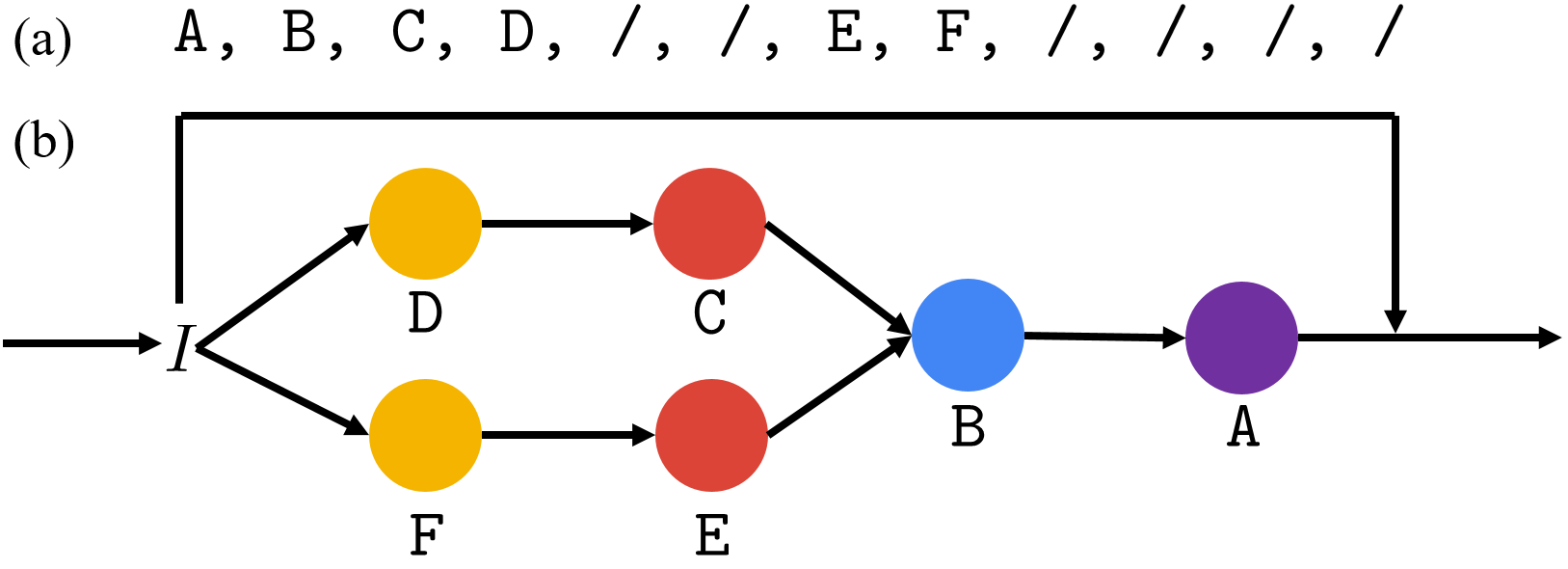}
    \end{center}
    \beforecaption
    \caption{An example computation graph of \ac{drt}. (a) Given the serialized $n$-ary tree representation (pre-order traversal with \texttt{/} denoting end-of-branch), (b) a tree-structured computation graph is dynamically built. The input features are wired from bottom-up following the tree structure. The final output is the sum with the input, forming a residual module.}
    \label{fig:drt}
    \aftercaption
\end{figure}

The image space of \ac{rpm} is inherently structured and could be described using a symbolic language, as shown in~\cite{carpenter1990one,lovett2017modeling,lovett2010structure,lovett2009solving,raven1938raven}. To capture this characteristic and further improve the model performance on \ac{rpm}, we propose a simple tree-structure neural module called \acf{drt} that operates on the joint space of image understanding and structure reasoning. An example of \ac{drt} is shown in Figure~\ref{fig:drt}.

In the \ac{drt}, given a sentence $S$ sampled from the \ac{asig}, usually represented as a serialized $n$-ary tree, we could first recover the tree structure. Note that the tree is \textbf{dynamically} generated following the sentence $S$, and each node in the tree comes with a label. With a structured tree representation ready, we could now consider assigning a neural computation operator to each tree node, similar to Tree-LSTM~\cite{tai2015improved}. To further simplify computation, we replace the LSTM cell~\cite{hochreiter1997long} with a ReLU-activated~\cite{nair2010rectified} fully-connected layer $f$. In this way, nodes with a single child (leaf nodes or OR-production nodes) update the input features by
\begin{equation}\label{eqn:or}
    I = \text{ReLU}(f([I, w_n])),
\end{equation}
where $[\cdot, \cdot]$ is the concatenation operation, $I$ denotes the input features, and $w_n$ the distributed representations of the node's label~\cite{mikolov2013distributed,pennington2014glove}. Nodes with multiple children (AND-production nodes) update input features by
\begin{equation}\label{eqn:and}
    I = \text{ReLU}\left(f\left(\left[\sum_c I_c, w_n\right]\right)\right),
\end{equation}
where $I_c$ denotes the features from its child $c$. 

In summary, features from the lower layers are fed into the leaf nodes of \ac{drt}, gradually updated by Equation~\ref{eqn:or} and Equation~\ref{eqn:and} from bottom-up following the tree structure, and output to higher-level layers.

Inspired by~\cite{he2016deep}, we make \ac{drt} a \textbf{residual} module by adding the input and output of \ac{drt} together, hence the name \acf{drt}
\begin{equation}
    I = \text{DRT}(I, S) + I.
\end{equation}

\section{Experiments}\label{sec:experiments}

\subsection{Computer Vision Models}

We adopt several representative models suitable for \ac{rpm} and test their performances on RAVEN~\cite{barrett2018measuring,he2016deep,krizhevsky2012imagenet,xingjian2015convolutional}. In summary, we test a simple sequential learning model (LSTM), a CNN backbone with an MLP head (CNN), a ResNet-based~\cite{he2016deep} image classifier (ResNet), the recent relational \ac{wren}~\cite{barrett2018measuring}, and all these models augmented with the proposed \ac{drt}. 

\paragraph{LSTM}
The partially sequential nature of the \ac{rpm} problem inspires us to borrow the power of sequential learning. Similar to ConvLSTM~\cite{xingjian2015convolutional}, we feed each image feature extracted by a CNN into an LSTM network sequentially and pass the last hidden feature into a two-layer MLP to predict the final answer. In the \ac{drt}-augmented LSTM, \ie, LSTM-\ac{drt}, we feed features of each image to a shared \ac{drt} before the final LSTM.

\paragraph{CNN}
We test a neural network model used in Hoshen and Werman~\cite{hoshen2017iq}. In this model, a four-layer CNN for image feature extraction is connected to a two-layer MLP with a softmax layer to classify the answer. The CNN is interleaved with batch normalization~\cite{ioffe2015batch} and ReLU non-linearity~\cite{nair2010rectified}. Random dropout~\cite{srivastava2014dropout} is applied at the penultimate layer of MLP. In CNN-\ac{drt}, image features are passed to \ac{drt} before MLP.

\paragraph{ResNet}
Due to its surprising effectiveness in image feature extraction, we replace the feature extraction backbone in CNN with a ResNet~\cite{he2016deep} in this model. We use a publicly available ResNet implementation, and the model is randomly initialized without pre-training. After testing several ResNet variants, we choose ResNet-18 for its good performance. The \ac{drt} extension and the training strategy are similar to those used in the CNN model.

\paragraph{\ac{wren}} 
We follow the original paper~\cite{barrett2018measuring} in implementing the \ac{wren}. In this model, we first extract image features by a CNN. Each answer feature is then composed with each context image feature to form a set of ordered pairs. The order pairs are further fed to an MLP and summed. Finally, a softmax layer takes features from each candidate answer and makes a prediction. In \ac{wren}-\ac{drt}, we apply \ac{drt} on the extracted image features before the relational module.

For all \ac{drt} extensions, nodes in the same level share parameters and the representations for nodes' labels are fixed after initialization from corresponding $300$-dimension GloVe vectors~\cite{pennington2014glove}. Sentences used for assembling \ac{drt} could be either retrieved or learned by an encoder-decoder. Here we report results using retrieval. 

\begin{table*}[ht]
    \begin{center}
        \caption{Testing accuracy of each model against human subjects and the solver. Acc denotes the mean accuracy of each model, while other columns show model accuracy on different figure configurations. \texttt{L-R} denotes \texttt{Left-Right}, \texttt{U-D} denotes \texttt{Up-Down}, \texttt{O-IC} denotes \texttt{Out-InCenter}, and \texttt{O-IG} denotes \texttt{Out-InGrid}. $^\star$Note that the perfect solver has access to rule operations and searches on the symbolic problem representation.}\label{tbl:plain_training}
        \begin{tabular}{l c c c c c c c c}
            \hline
            Method  & Acc & \texttt{Center} & \texttt{2x2Grid} & \texttt{3x3Grid} & \texttt{L-R} & \texttt{U-D} & \texttt{O-IC} & \texttt{O-IG} \\
            \hline\hline
            LSTM                & 13.07\% & 13.19\% & 14.13\% & 13.69\% & 12.84\% & 12.35\% & 12.15\% & 12.99\% \\
            \ac{wren}           & 14.69\% & 13.09\% & 28.62\% & 28.27\% & 7.49\%  & 6.34\%  & 8.38\%  & 10.56\% \\
            CNN                 & 36.97\% & 33.58\% & 30.30\% & 33.53\% & 39.43\% & 41.26\% & 43.20\% & 37.54\% \\
            ResNet              & 53.43\% & 52.82\% & 41.86\% & 44.29\% & 58.77\% & 60.16\% & 63.19\% & 53.12\% \\
            \hline
            LSTM+DRT            & 13.96\% & 14.29\% & 15.08\% & 14.09\% & 13.79\% & 13.24\% & 13.99\% & 13.29\% \\
            \ac{wren}+DRT       & 15.02\% & 15.38\% & 23.26\% & 29.51\% & 6.99\%  & 8.43\%  & 8.93\%  & 12.35\% \\
            CNN+DRT             & 39.42\% & 37.30\% & 30.06\% & 34.57\% & 45.49\% & 45.54\% & 45.93\% & 37.54\% \\
            \textbf{ResNet+DRT} & \textbf{59.56\%} & \textbf{58.08\%} & \textbf{46.53\%} & \textbf{50.40\%} & \textbf{65.82\%} & \textbf{67.11\%} & \textbf{69.09\%} & \textbf{60.11\%} \\
            \hline
            Human               & 84.41\% & 95.45\% & 81.82\% & 79.55\% & 86.36\% & 81.81\% & 86.36\% & 81.81\% \\
            Solver$^\star$      & 100\%   & 100\%   & 100\%   & 100\%   & 100\%   & 100\%   & 100\%   & 100\%   \\
            \hline
        \end{tabular}
        \aftertable
    \end{center}
\end{table*}

\subsection{Experimental Setup}

We split the RAVEN dataset into three parts, $6$ folds for training, $2$ folds for validation, and $2$ folds for testing. We tune hyper-parameters on the validation set and report the model accuracy on the test set. For loss design, we treat the problem as a classification task and train all models with the cross-entropy loss. All the models are implemented in PyTorch~\cite{paszke2017automatic} and trained with ADAM~\cite{kingma2014adam} before early stopping or a maximum number of epochs is reached.

\subsection{Performance Analysis}

Table \ref{tbl:plain_training} shows the testing accuracy of each model trained on RAVEN, against the human performance and the heuristics-based solver. Neither human subjects nor the solver experiences an intensive training session, and the solver has access to the rule operations and searches the answer based on a symbolic representation of the problem. In contrast, all the computer vision models go over an extensive training session, but only on the training set.

In general, human subjects produce better testing accuracy on problems with simple figure configurations such as \texttt{Center}, while human performance reasonably deteriorates on problem instances with more objects such as \texttt{2x2Grid} and \texttt{3x3Grid}. Two interesting observations:
\begin{enumerate}[leftmargin=*,noitemsep,nolistsep]
    \item For figure configurations with multiple components, although each component in \texttt{Left-Right}, \texttt{Up-Down}, and \texttt{Out-InCenter} has only one object, making the reasoning similar to \texttt{Center} except that the two components are independent, human subjects become less accurate in selecting the correct answer.
    \item Even if \texttt{Up-Down} could be regarded as a simple transpose of \texttt{Left-Right}, there exists some notable difference. Such effect is also implied by the ``inversion effects'' in cognition; for instance, inversion disrupts face perception, particularly sensitivity to spatial relations~\cite{crookes2009early,le2001neuroperception}.
\end{enumerate}

In terms of model performance, a counter-intuitive result is: computer vision systems do not achieve the best accuracy across all other configurations in the seemingly easiest figure configuration for human subjects (\texttt{Center}). We further realize that the LSTM model and the \ac{wren} model perform only slightly better than random guess (12.5\%). Such results contradicting to~\cite{barrett2018measuring} might be attributed to the diverse figure configurations in RAVEN. Unlike LSTM whose accuracy across different configurations is more or less uniform, \ac{wren} achieves higher accuracy on configurations consisting of multiple randomly distributed objects (\texttt{2x2Grid} and \texttt{3x3Grid}), with drastically degrading performance in configurations consisting of independent image components. This suggests \ac{wren} is biased to grid-like configurations (majority of PGM) but not others that require compositional reasoning (as in RAVEN). In contrast, a simple CNN model with MLP doubles the performance of \ac{wren} on RAVEN, with a tripled performance if the backbone is ResNet-18.

We observe a consistent performance improvement across different models after incorporating \ac{drt}, suggesting the effectiveness of the structure information in this visual reasoning problem. While the performance boost is only marginal in LSTM and \ac{wren}, we notice a marked accuracy increase in the CNN- and ResNet-based models (6.63\% and 16.58\% relative increase respectively). However, the performance gap between artificial vision systems and humans are still significant (up to 37\% in \texttt{2x2Grid}), calling for further research to bridge the gap.

\subsection{Effects of Auxiliary Training}

Barrett \etal~\cite{barrett2018measuring} mentioned that training \ac{wren} with a fine-tuned auxiliary task could further give the model a 10\% performance improvement. We also test the influence of auxiliary training on RAVEN. First, we test the effects of an auxiliary task to classify the rules and attributes on \ac{wren} and our best performing model ResNet+\ac{drt}. The setting is similar to~\cite{barrett2018measuring}, where we perform an OR operation on a set of multi-hot vectors describing the rules and the attributes they apply to. The model is then tasked to both correctly find the answer and classify the rule set with its governing attributes. The final loss becomes 
\begin{equation}
    \mathcal{L}_{\text{total}} = \mathcal{L}_{\text{target}} + \beta \mathcal{L}_{\text{rule}},
\end{equation}
where $\mathcal{L}_\text{{target}}$ denotes the cross-entropy loss for the answer, $\mathcal{L}_{\text{rule}}$ the multi-label classification loss for the rule set, and $\beta$ the balancing factor. We observe no performance change on \ac{wren} but a serious performance downgrade on ResNet+\ac{drt} (from 59.56\% to 20.71\%).

Since RAVEN comes with structure annotations, we further ask whether adding a structure prediction loss could help the model improve performance. To this end, we cast the experiment in a similar setting where we design a multi-hot vector describing the structure of each problem instance and train the model to minimize
\begin{equation}
    \mathcal{L}_{\text{total}} = \mathcal{L}_{\text{target}} + \alpha \mathcal{L}_{\text{struct}},
\end{equation}
where $\mathcal{L}_{\text{struct}}$ denotes the multi-label classification loss for the problem structure, and $\alpha$ the balancing factor. In this experiment, we observe a slight performance decrease in ResNet+\ac{drt} (from 59.56\% to 56.86\%). A similar effect is noticed on \ac{wren} (from 14.69\% to 12.58\%).

\setstretch{0.98}

\subsection{Test on Generalization}

One interesting question we would like to ask is how a model trained well on one figure configuration performs on another similar figure configuration. This could be a measure of models' generalizability and compositional reasoning ability. Fortunately, RAVEN naturally provides us with a test bed. To do this, we first identify several related configuration regimes: 
\begin{itemize}[leftmargin=*,noitemsep,nolistsep]
    \item Train on \texttt{Center} and test on \texttt{Left-Right}, \texttt{Up-Down}, and \texttt{Out-InCenter}. This setting directly challenges the compositional reasoning ability of the model as it requires the model to generalize the rules learned in a single-component configuration to configurations with multiple independent but similar components.
    
    \item Train on \texttt{Left-Right} and test on \texttt{Up-Down}, and vice-versa. Note that for \texttt{Left-Right} and \texttt{Up-Down}, one could be regarded as a transpose of another. Thus, the test could measure whether the model simply memorizes the pattern in one configuration.
    
    \item Train on \texttt{2x2Grid} and test on \texttt{3x3Grid}, and vice-versa. Both configurations involve multi-object interactions. Therefore the test could measure the generalization when the number of objects changes.
\end{itemize}
The following results are all reported using the best performing model, \ie, ResNet+\ac{drt}.

\begin{table}[ht]
    \begin{center}
        \caption{Generalization test. The model is trained on \texttt{Center} and tested on three other configurations.}\label{tbl:test_gen_ctr}
        \resizebox{\linewidth}{!}{
        \begin{tabular}{c c c c}
            \hline
            \texttt{Center} & \texttt{Left-Right} & \texttt{Up-Down} & \texttt{Out-InCenter} \\
            \hline\hline
            51.87\%         & 40.03\%             & 35.46\%          & 38.84\%            \\ 
            \hline
        \end{tabular}}
        \aftertable
    \end{center}
\end{table}

\begin{table}[ht]
    \begin{center}
        \beforetable
        \caption{Generalization test. The row shows configurations the model is trained on and the column the model is tested on.}\label{tbl:test_gen_two}
        \begin{tabular}{l c c}
            \hline
                                & \texttt{Left-Right} & \texttt{Up-Down} \\
            \hline\hline
            \texttt{Left-Right} & 41.07\%             & 38.10\%          \\
            \texttt{Up-Down}    & 39.48\%             & 43.60\%          \\
            \hline
        \end{tabular}
        \aftertable
    \end{center}
\end{table}

\begin{table}[ht]
    \begin{center}
        \beforetable
        \caption{Generalization test. The row shows configurations the model is trained on and the column the model is tested on.}\label{tbl:test_gen_multi}
        \begin{tabular}{l c c}
            \hline
                             & \texttt{2x2Grid} & \texttt{3x3Grid} \\
            \hline\hline
            \texttt{2x2Grid} & 40.93\%          & 38.69\%       \\
            \texttt{3x3Grid} & 39.14\%          & 43.72\%       \\
            \hline
        \end{tabular}
        \aftertable
    \end{center}
\end{table}

Table~\ref{tbl:test_gen_ctr},~\ref{tbl:test_gen_two} and~\ref{tbl:test_gen_multi} show the result of our model generalization test. We observe:
\begin{itemize}[leftmargin=*,noitemsep,nolistsep]
    \item The model dedicated to a single figure configuration does not achieve better test accuracy than one trained on all configurations together. This effect justifies the importance of the diversity of RAVEN, showing that increasing the number of figure configurations could actually improve the model performance.
    \item Table~\ref{tbl:test_gen_ctr} also implies that a certain level of compositional reasoning, though weak, exists in the model, as the three other configurations could be regarded as a multi-component composition of \texttt{Center}.
    \item In Table \ref{tbl:test_gen_two}, we observe no major differences in terms of test accuracy. This suggests that the model could successfully transfer the knowledge learned in a scenario to a very similar counterpart, when one configuration is the transpose of another.
    \item From Table \ref{tbl:test_gen_multi}, we notice that the model trained on \texttt{3x3Grid} could generalize to \texttt{2x2Grid} with only minor difference from the one dedicated to \texttt{2x2Grid}. This could be attributed to the fact that in the \texttt{3x3Grid} configuration, there could be instances with object distribution similar to that in \texttt{2x2Grid}, but not vice versa.
\end{itemize}

\section{Conclusion}\label{sec:conclusion}

We present a new dataset for Relational and Analogical Visual Reasoning in the context of \acf{rpm}, called RAVEN. Unlike previous work, we apply a systematic and structured tool, \ie, \acf{asig}, to generate the dataset, such that every problem instance comes with rich annotations. This tool also makes RAVEN diverse and easily extendable. One distinguishing feature that tells apart RAVEN from other work is the introduction of the structure. We also recruit quality human subjects to benchmark human performance on the RAVEN dataset. These aspects fill two important missing points in previous works.

We further propose a novel neural module called \acf{drt} that leverages the structure annotations for each problem. Extensive experiments show that models augmented with \ac{drt} enjoy consistent performance improvement, suggesting the effectiveness of using structure information in solving \ac{rpm}. However, the difference between machine algorithms and humans clearly manifests itself in the notable performance gap, even in an unfair situation where machines experience an intensive training session while humans do not. We also realize that auxiliary tasks do not help performance on RAVEN. The generalization test shows the importance of diversity of the dataset, and also indicates current computer vision methods do exhibit a certain level of reasoning ability, though weak.

The entire work still leaves us many mysteries. Humans seem to apply a combination of the top-down and bottom-up method in solving \ac{rpm}. How could we incorporate this into a model? What is the correct way of formulating visual reasoning? Is it model fitting? Is deep learning the ultimate way to visual reasoning? If not, how could we revise the models? If yes, how could we improve the models? 

Finally, we hope these unresolved questions would call for attention into this challenging problem.

{\textbf{Acknowledgement:}
The authors thank Prof. Ying Nian Wu and Prof. Hongjing Lu at UCLA Statistics Department for helpful discussions. The work reported herein was supported by DARPA XAI grant N66001-17-2-4029, ONR MURI grant N00014-16-1-2007, ARO grant W911NF-18-1-0296, and a NVIDIA GPU donation.}

\newpage

\bibliographystyle{ieee}
\bibliography{bib}

\end{document}